\documentclass[10pt,twocolumn,letterpaper]{article}

\usepackage{wacv}
\usepackage{times}
\usepackage{epsfig}
\usepackage{graphicx}
\usepackage{caption}
\usepackage{subcaption}
\usepackage{amsmath}
\usepackage{amssymb}

\newcommand\Mark[1]{\textsuperscript#1}

\wacvfinalcopy 

\def\wacvPaperID{79} 

\ifwacvfinal\fi
\begin{document}

\title{Dynamic Belief Fusion for Object Detection}

\author{
Hyungtae Lee\Mark{1}, Heesung Kwon\Mark{1}, Ryan M. Robinson\Mark{1}, William D. Nothwang\Mark{1}, and Amar M. Marathe\Mark{2}\\
\Mark{1}SEDD, U.S. Army Research Laboratory, Adelphi, MD, USA\\
\Mark{2}HRED, U.S. Army Research Laboratory, Aberdeen, MD, USA\\
{\tt\small htlee@umd.edu}\\
{\tt\small \{heesung.kwon.civ,ryan.robinson14.ctr,william.d.nothwang,amar.marathe.civ\}@mail.mil}
}



\maketitle
\ifwacvfinal\thispagestyle{empty}\fi

\begin{abstract}
A novel approach for the fusion of heterogeneous object detection methods is proposed.  In order to effectively integrate the outputs of multiple detectors, the level of ambiguity in each individual detection score is estimated using the precision/recall relationship of the corresponding detector.  The main contribution of the proposed work is a novel fusion method, called Dynamic Belief Fusion (DBF), which dynamically assigns probabilities to hypotheses (target, non-target, intermediate state (target or non-target)) based on confidence levels in the detection results conditioned on the prior performance of individual detectors.  In DBF, a joint basic probability assignment, optimally fusing information from all detectors, is determined by the Dempster's combination rule, and is easily reduced to a single fused detection score.  Experiments on ARL and PASCAL VOC 07 datasets demonstrate that the detection accuracy of DBF is considerably greater than conventional fusion approaches as well as individual detectors used for the fusion.
\end{abstract}

\vspace{-0.35cm}
\section{Introduction}


\begin{figure}[t]
    \begin{center}
    \includegraphics[trim=15mm 25mm 15mm 10mm,width=0.46\textwidth]{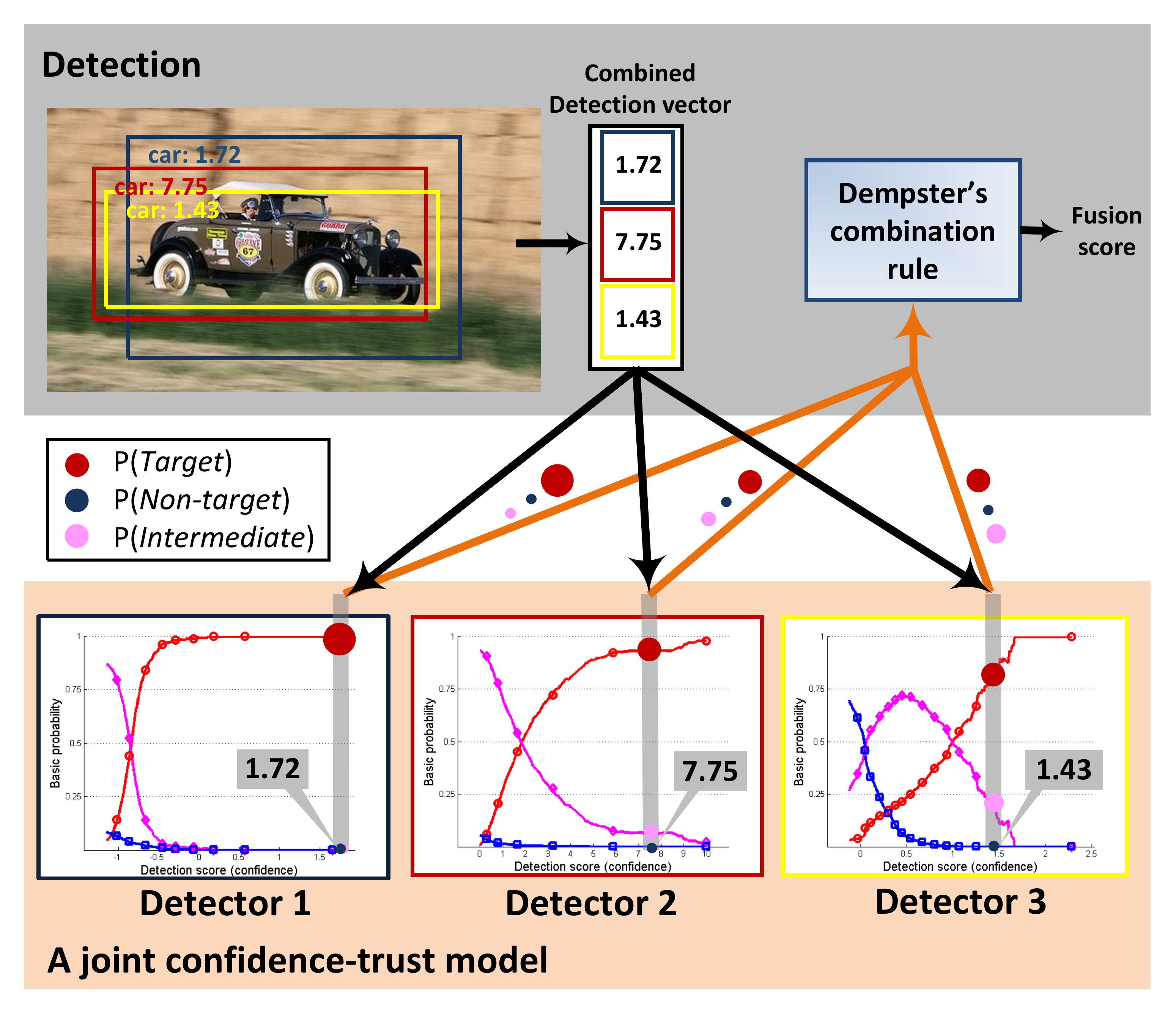}
    \end{center}
    \caption{{\bf Dynamic Belief Fusion:} Three detectors colored by blue, red, and yellow detect a car in a given image, as shown on the top of the figure.  A combined detection vector is constructed by collecting detection scores whose windows overlap each other.  For each detector, the basic probabilities of target (red), non-target (blue), and intermediate state (target or non-target) (pink), shown at the bottom, which dynamically vary as a function of detection score in conjunction with the trust model representing prior information of each detector, are assigned.  (In each plot, the circle radius represents the magnitude of basic probability assignment.)  Dempster's combination rule combines the basic probabilities of each detector and returns a fused confidence score.  We call this fusion precess as Dynamic Belief Fusion (DBF).}
    \label{fig:fusion}
    \vspace{-0.30cm}
\end{figure}

Current methods for fusing multiple object detectors are often specific to a subset of detectors with shared features~\cite{BFernandoCVPR12,XLanCVPR14,PNatarajanCVPR12,HWangCVPR13}.  However, the field of object detection is undergoing a state of rapid advancement~\cite{NDalalCVPR05,PFelzenszwalbPAMI10,TMalisiewiczICCV11}.  Many detection algorithms, and hence, feature-specific fusion algorithms, are quickly becoming obsolete.  There is an increasing need for fusion methods that can combine object detection algorithms regardless of their structure.  
One effective solution in this case can be late fusion, a process which conditions the ``trust'' in individual detector outputs on their prior performance, and then intelligently combines the trust-weighted outputs.

Several approaches to late fusion exist, including Bayesian fusion and Dempster-Shafer Theory (DST) fusion.  However, Bayesian fusion typically does not yield significant improvements in performance (see Table~\ref{tab:map_pascal}) due to its inherent characteristics.   Bayesian fusion handles uncertainty in a detector's output by associating a probability to each hypothesis (e.g. 30\% chance of \emph{target} and 70\% chance of \emph{non-target}); however, the Bayesian approach does not indicate the level of trust to be placed in the probability assignments themselves.  Belief theory, a component of DST developed by G. Shafer~\cite{GShaferPrinceton76}, takes a step in the right direction to address the ambiguity in detector quality through its use of compound hypotheses.  In considering two hypotheses, \emph{target} and \emph{non-target}, Shafer's belief theory assigns probability to the information that directly supports the \emph{target} and \emph{non-target} hypotheses, and also instantiates an intermediate state, \emph{target OR non-target}, with its own probability quantifying the level of ambiguity that makes either hypothesis plausible.  In this manner, a detector output with a high level of ambiguity can be ignored/down-weighted in favor of a more trustworthy, low-ambiguity detector output.  However, assigning these belief probabilities is not a trivial task, and choice of assignment method is critical to fusion performance.

We propose a novel approach called Dynamic Belief Fusion (DBF), which assigns probability to hypotheses dynamically under the framework of DST.  In this approach, trust in an information source is characterized as a continuous function of its output by assigning a corresponding set of probabilities to each output value.  The DBF process is partly illustrated in Figure~\ref{fig:fusion}, in which three heterogeneous detectors generate scores for a target candidate window.  Similar to other late fusion methods, these scores are cross-referenced with the detectors' trust models to obtain a set of probability assignments, essentially re-weighting the outputs of each detector.  The probabilities from multiple detectors are then combined into a single fused detection score via Dempster's combination rule~\cite{ADempsterAMS67}.

In order to generate continuous probability assignments (also known as belief functions or trust models) for \emph{target}, \emph{non-target}, and \emph{intermediate state} hypotheses in the context of object detection, we employ the precision-recall (PR) model of each detector in a validation step.  Specifically, in order to compute the probability assigned to the intermediate state, we devise the notion of a \emph{best possible} detector, a theoretical detector trained over a limited number of images that can generate best detection performance possible close to a theotical limit.  We estimate the PR curve of the best possible detector, and treat the difference in precision between any individual detector and the best possible detector as the ambiguity in the decision of the individual detector and thus, assign it to the probability of the individual detector's intermediate state.

Our contributions are summarized as following:
\begin{enumerate}
	\item We introduce a novel late fusion framework by optimaly modeling joint relationships between a priori and current information of individual detectors.  The proposed fusion approach can robustly extract complementary information from multiple disparate detection approaches consistently generating superior performance over the best individual detector.  We believe these results, as well as the clear improvement over existing late fusion algorithms, will inspire greater efforts along the lines of late fusion research.
	\item Our novel approach computes the probabilities dynamically, which are assigned to all constituent hypotheses, including an intermediate state (target or non-target), by optimally linking the current confidence levels in detection (i.e. detection scores) to the precision-recall relationships estimated from a validation set as prior information.
\end{enumerate}


The proposed DBF method is evaluated using ARL~\cite{JTouryanNeuroprosthetics14} and PASCAL VOC 07~\cite{pascal-voc-2007} image sets.  DBF is compared to other well-known fusion methods.  In these experiments, DBF outperforms all individual computer vision methods, as well as other fusion methods.

\section{Related Works}

We can split the literature concerning the fusion of multiple heterogeneous information sources into two categories: (i) building a joint model by integrating multiple approaches and (ii) fusing the output of multiple approaches.  

Kwon and Lee proposed two approaches integrating multiple sample-based tracking approaches using an interactive Markov Chain Monte Carlo (iMCMC) framework~\cite{JKwonCVPR10} and using sampling in tracker space modeled by Markov Chain Monte Carlo (MCMC) method~\cite{JKwonICCV11}, respectively.  We et al.~\cite{SWuCVPR14} introduced an approach combining detectors of different modalities (concept, text, speech) by using relationships among the modes in the event detection.  However, in general, modeling the dependencies in fusion among multiple approaches built on different principles is infeasible.

In the case where modeling dependency among multiple approaches is not possible, fusion can be performed over their outputs (late fusion).  Bailer et al.~\cite{CBailerECCV14} introduced a fusion framework in the target tracking paradigm.  They collected trajectories from multiple tracking algorithms and computed one fused trajectory in order to improve accuracy, trajectory continuity, and smoothness.  However, since temporal information obtained from trajectory cannot be applicable in the object detection task, a different fusion framework is necessary for our problem.  Kim et al.~\cite{THKimICCV13} and Liu et al.~\cite{DLiuCVPR13} used weighted-sum (WS) methods to fuse multiple types of data for object detection.  Their WS method learns weights in a manner that estimates trust in multiple data sources.  However, since weights optimization is usually performed to maximize distance between positive and negative samples, like Bayesian fusion, WS does not provide a way to indicate ambiguity between the positives and negatives, which degrades fusion performance, as previously mentioned.  The works of Ma and Yuen~\cite{AJMaECCV12} and Liu et al.~\cite{CLiuPAMI11} employing Bayesian fusion also shows limited performance. 

In order to improve upon these late fusion results, we introduce DBF, a general fusion framework for object detection, employing DST to interpret and leverage ambiguity more completely.  In Section~\ref{sec:exp}, experiments demonstrate the prominent performance of our proposed approach against WS and Bayesian fusion, as well as other existing methods.

\section{Dempster-Shafer Theory}
\label{sec:DST}


In this section, we detail components of Dempster-Shafer theory (DST) which form the foundation of our proposed DBF method.  Dempster-Shafer theory~\cite{ADempsterAMS67,GShaferPrinceton76} is based on Shafer's belief theory~\cite{GShaferPrinceton76} that obtains a degree of belief for a hypothesis by combining evidences from probabilities of related hypotheses.  DST combines such beliefs from multiple independent sources using a method develped by A. Dempster.

\subsection{Shafer's Belief Theory}

Let $X$ be a universal set consisting of $M$ exhaustive and mutually exclusive hypotheses, i.e. $X = \{1, 2, \cdots , M\}$.  The power set $2^X$ is the set of all subsets of $X$.  Basic probability in the range $[0~1]$ is assigned to each element of the power set $2^X$.  A function defined as $m:2^X \rightarrow [0~1]$ is called a basic probability assignment (BPA).  Subsets consisting of compound hypotheses in $X$ represent ambiguity among the constituent hypotheses; the BPA given to the subset measures the level of ambiguity.  A BPA has two properties; (i) $m(\emptyset) = 0$ (the mass of the empty set is zero) and (ii) $\sum_{A\in 2^X}{m(A)=1}$ (the BPA values of the members of the power set sum to one).

From the BPAs, the belief function $bel(A)$ for a set $A$ can be defined as the sum of all masses which are subsets of the set of interest:
\begin{equation}
bel(A)=\sum_{B|B\subseteq A}{m(B)}.\label{eq:belief}
\end{equation}
Belief represents the information in direct support of $A$.

\subsection{Dempster's Combination Rule}

Dempster's combination rule can be applied to calculate a \emph{joint BPA} from separate BPAs.  Under the condition that the evidence from each pair is independent of the other, Dempster's combination rule defines a joint BPA $m_f = m_1 \oplus m_2$, which represents the combined effect of $m_1$ and $m_2$, i.e.,
\begin{equation}
m_f(A)=m_1 \oplus m_2 (A)=\frac{1}{N}\sum_{X\cap Y=A,~A\neq\emptyset}{m_1 (X)m_2 (Y)},
\end{equation}
where $N=\sum_{X\cap Y\neq\emptyset}{m_1 (X) m_2 (Y)}$ and $X$ and $Y$ are subsets of $2^X$.  $N$ is a measure of the amount of any mass whose common evidence is not the null set.  Dempster's rule can be extended for multiple pieces of evidence (e.g., multiple detectors) using the associative and commutative properties of BPAs (i.e. $m_f = m_1\oplus m_2 \oplus\cdots\oplus m_K.$) with the following formula:
\begin{equation}
m_f(A)=\frac{1}{N}\sum_{X_1\cap X_2\cap \cdots \cap  X_K=A}{\prod_{i=1}^K{m_i (X_i)}},
\label{eq:comb_rule}
\end{equation}
where $N=\sum_{X_1 \cap \cdots \cap X_K\neq\emptyset}{\prod_{i=1}^{K}{m_i (X_i)}}$.

\section{The Proposed Fusion Approach}

\subsection{Overview of the Fusion of Detectors}

\begin{figure*}[t]
    \begin{center}
    \includegraphics[trim = 10mm 65mm 10mm 65mm,width=\textwidth]{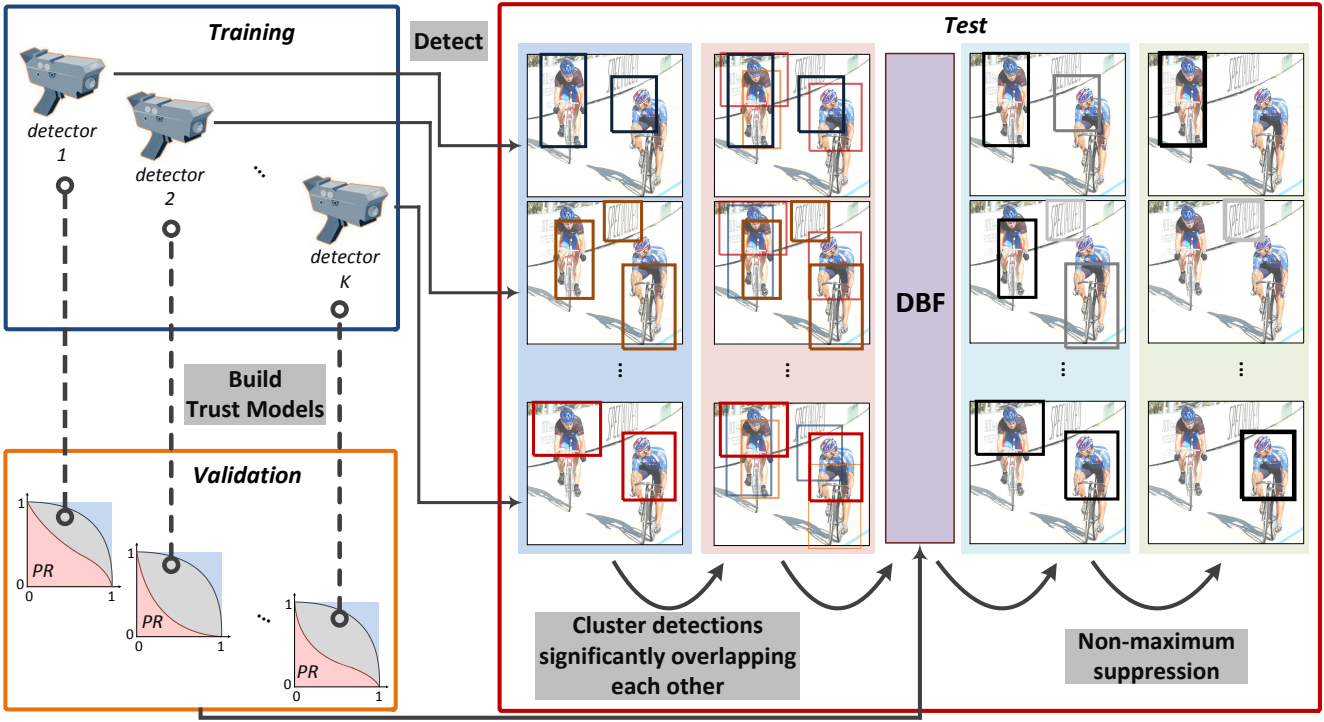}
    \end{center}
    \caption{Flow diagram of the proposed fusion algorithm.  In the 4th and 5th columns (right side of ``DBF'') of the ``test'' step, darker windows indicate higher confidence.}
    \label{fig:flow_chart}
\end{figure*}

The proposed fusion of object detectors is performed in three steps. (i) Individual detectors are trained on a training set. (ii) A trust model as prior information for each individual detector is constructed by calculating precision/recall (PR) relationships for individual detectors using detection scores obtained from the validation set and ground truth information.  (iii) For testing, given detection scores from the test set, the precision/recall relationships along with current detection scores are used to assign probabilities to all hypotheses of detections from individual detectors.  The probabilities of individual hypotheses for a given observation are estimated by adaptively linking the current detection scores to the previously estimated PR models.  Joint exploitation of the current detection scores and the past PR model is used to estimate trustworthiness of the current detections in conjunction with the past performance of individual detectors.  The estimated probabilities for individual hypotheses are separately fused over different detection approaches using Equation~\ref{eq:comb_rule}.   Figure~\ref{fig:flow_chart} illustrates the fusion process of the proposed DBF algorithm.  Details of the proposed fusion process are as follows.

$ $

\noindent {\bf Building a trust model as a prior performance model for individual detectors (validation):} Detectors are applied to validation images in a scanning window fashion and to search for potential objects of interest.  To construct the trust model of each detector we first estimate the PR relationships of all the detectors.  In building the PR model, all detection windows are labeled as \emph{true}  or \emph{false positive} or \emph{undecided} by comparing them with ground truth (annotated windows containing the objects of interest).  Any detection
window that has an intersection-over-union overlap (PASCAL VOC criteria~\cite{pascal-voc-2007}) of greater than 0.5 with a groundtruth window is assigned \emph{true positive}.  If there is no overlap between a detection window and a groundtruth window, the detection is assigned \emph{false positive}.  The remaining detections are labeled \emph{undecided}.  The PR model is constructed using the labeled detection results.


$ $

\noindent {\bf Constructing a combined detection vector from detection windows for fusion (test):} Let  $d_j^i, i = 1, 2, \cdots, K, j = 1, 2, \cdots, W_i$ be the $j^{th}$ window of the $i^{th}$ detector associated with detection score $c_j^i$.  $K$ is the number of detectors and $W_i$ is the number of detection windows of the $i^{th}$ detector.  For each detection from all the detectors given a test image, we collect the detection windows from the remaining detectors that significantly overlap the subject detection window (see the second column of the test phase in Figure~\ref{fig:flow_chart}).  Two detections $d_j^i$ and $d_l^k$, $i\neq k$ are considered significantly overlapping if the intersection-over-union overlap of their windows is greater than 0.5.  A $K$-dimensional detection vector ${\bf c} = [c_{j_1}^1~c_{j_2}^2~\cdots~c_{j_K}^K]$ is then constructed, consisting of the score of the subject detection window and those of the overlapped windows from other detectors.  If multiple windows from the same detector overlap the subject detection window, the window with the maximum detection score between them is used. If no overlaps exist for a particular detector, the corresponding element of the combined detection vector is filled by a value of negative infinity to ignore the influence of the detector in fusion.

$ $

\noindent {\bf Fusing detection windows (test):} Fusion is performed over the combined detection vector using DBF.  Details of DBF are described in Section~\ref{ssec:DDST}.  DBF dynamically assigns basic probabilities to the hypotheses of a given observation by adaptively mapping current detection scores to the PR model, which are fused over all the detectors by the Dempster's combination rule.  After rescoring all windows by applying DBF, non-maximum suppression is applied to merge windows whose intersection over union overlap is greater than 0.5.  The final output of the fusion procedure is a consolidated set of windows, each with a fused detection score.

\subsection{Dynamic Belief Fusion}
\label{ssec:DDST}

In binary object detection, the universal set $X$ is defined as $\{T,\neg T\}$ and thus its power set is expressed as $\{\emptyset, T,\neg T, \{T, \neg T\}\}$, where $T$ is a \emph{target} hypothesis and $\neg T = X - T$ is a \emph{non-target} hypothesis.  $\{T, \neg T\}$ in the power set represents detection ambiguity, denoted by $I$ ($\emph{intermediate state}$), which indicates that the subject observation could be either target or non-target.  We assign basic probabilities to all hypotheses based on prior detection performance of detectors.  We employ the precision-recall (PR) model 
to represent the prior information of individual detectors and compute basic probabilities of the hypotheses for a given observation.  Since the PR relationship is obtained by varying a threshold against detection scores, $c^i_j$, the basic probabilities being assigned dynamically change as $c^i_j$ changes.  Hence, we refer to this assignment as dynamic basic probability assignments.

\begin{figure}[t]
    \begin{center}
    \includegraphics[trim = 30mm 20mm 30mm 20mm,width=0.43\textwidth]{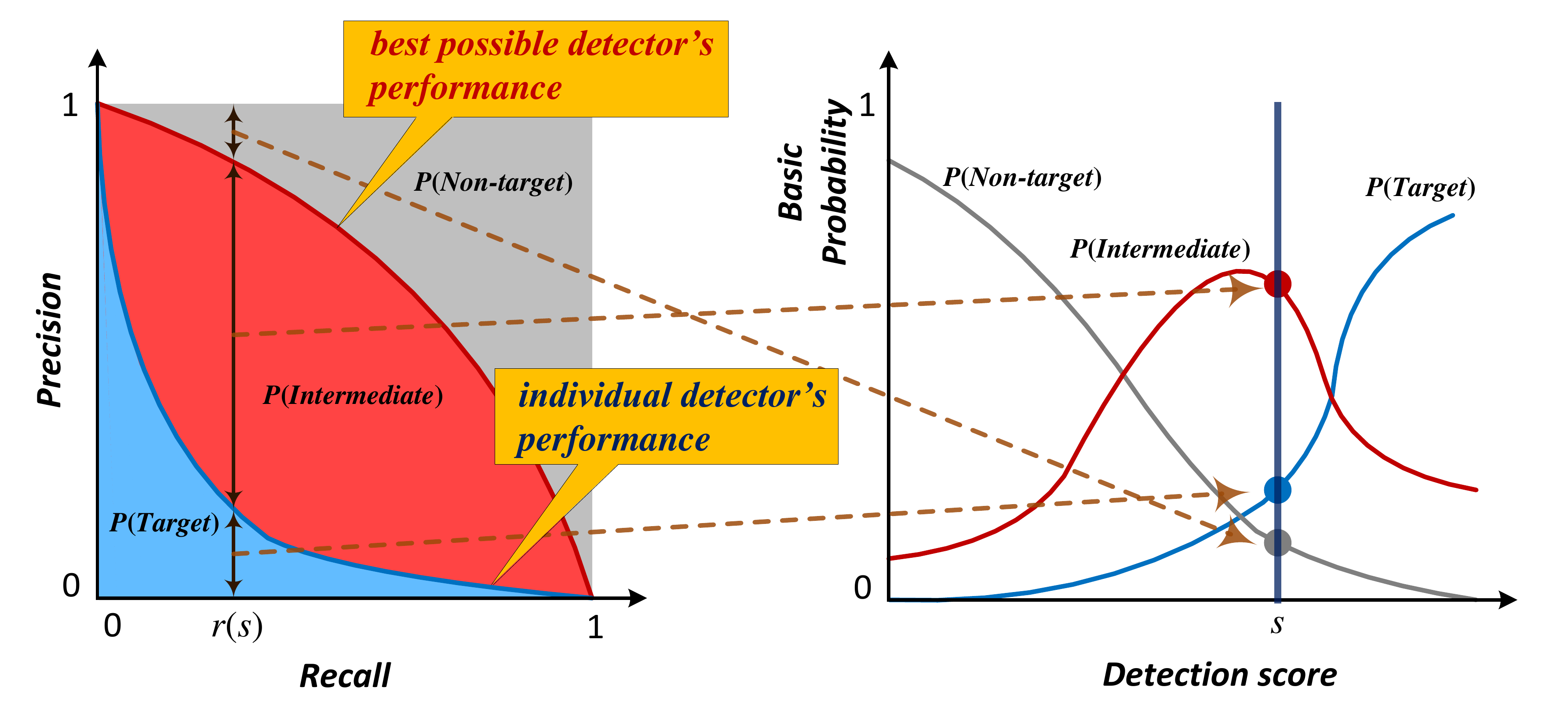}
    \end{center}
    \caption{{\bf Dynamic Basic Probability Assignment:} The left plot shows a precision/recall curve for an individual detector and a \emph{best possible} detector.  The rates of values along the precision axis corresponding to recall $r(s)$ are assigned as the basic probabilities to \emph{target}, \emph{non-target}, and \emph{intermediate state}, where $s$ is a detection score.  The right plot presents the basic probabilities with respect to a detection score, which converted from the PR curve.}
    \label{fig:pr_dba}
\end{figure}

In DBF as shown in Fig.~\ref{fig:pr_dba}, each element of the combined detection vector, $c^i_j$, is first mapped to the corresponding recall and the corresponding precision ($p$) is assigned as the basic probability of $\emph{target}$ hypothesis.  Then, $1-p$ needs to be split to account for two basic probabilities of $\emph{non-target}$ and $\emph{intermediate state}$ since it includes information about both hypotheses.  This is because precision is only defined for targets (not backgrounds).    Note that the recall of background (i.e., recall when ``positive'' refers to background) cannot be calculated because the number of backgrounds is close to infinite.

Since the split can not be achieved based solely on the given PR relationship, we introduce a theoretical best possible detector, whose performance can possibly achieve a level close to a theoretical limit.  Ideally, individual detectors can also achieve the same performance of the theoretical detector if they are provided with complete information about target and non-target.  In reality, individual detectors do not have complete information in training.  We treat the difference between the precision of an individual detector and that of the best possible detector as the detection ambiguity (i.e. the probability of the \emph{intermediate state}) caused by the lack of complete information in training.  In our work, the PR curve of the best possible detector, $\hat{p}_{bpd}$, is modeled as
\begin{equation}
\hat{p}_{bpd}(r) = 1 - r^n,
\label{eq:fn_bpd}
\end{equation}
where $r$ is recall.  This model is proposed because in general $\hat{p}_{bpd}$ should mimic the typical behavior of a highly accurate detector, a concave function approaching the top right corner of the plot such as the car detector in~\cite{PFelzenszwalbPAMI10}.  $m(I)$ is defined by $\hat{p}_{bpd} - p$ and the remaining fraction of precision $1-\hat{p}_{bpd}$ is assigned to $m(\neg T)$.  As $n$ approaches infinity, the best possible detector becomes the perfect detector (i.e. no false positives).  Dynamic basic probability assignment is shown in Figure~\ref{fig:pr_dba}.  Fusion of the detections from multiple individual detectors is achieved by computing fused basic probability assignments of $\emph{target}$ and $\emph{non-target}$ hypotheses, $m_f(T)$ and $m_f(\neg T)$, by Dempster's combination rule in Equation~\ref{eq:comb_rule}.  The overall fusion score is given by $s = bel(T) - bel(\neg T)$ where in our experiments, $bel(T)$ and $bel(\neg T)$ are actually $m_f(T)$ and $m_f(\neg T)$, respectively, according to Equation~\ref{eq:belief} since ${T}$ and $\neg T$ are sets of a single element.

\section{Experiments}
\label{sec:exp}

\subsection{Evaluation Setting}

\noindent	{\bf Individual Detectors:} Eight object detectors with unique detection structures whose codes are readily available online were selected: two SVM-based detectors incorporating both HOG~\cite{NDalalCVPR05} and Dense SIFT~\cite{CLiuPAMI11}, two Deformable-Part Model (DPM) with HOG~\cite{PFelzenszwalbPAMI10} and color attribute~\cite{FSKhanCVPR12}, Transductive Annotation by Graph (TAG)~\cite{JWangICML08}, exemplar SVM~\cite{TMalisiewiczICCV11}, and two CNN based detector (fine tuned CNN~\cite{MOquabCVPR14}, RCNN~\cite{RGirshickCVPR14}).  Given an image, the detection score indicates a degree of confidence about the decision.  The eight selected detectors use different feature extraction methods (e.g. HOG, dense SIFT, color attributes, and CNN features, etc.) and different principles of detecting objects of interest.

\noindent	{\bf Baselines (Fusion):} As a baseline, we used five approaches: Platt scaling~\cite{JPlattALMC99}, Weighted Sum (WS), Bayesian fusion, Local Expert Forest~\cite{JLiuECCV12}, and Detect2Rank~\cite{SKaraogluArxiv14}.  The Platt scaling learns a logistic regression model on the detection scores of true and false positive detections.  We applied Platt scaling to all the detectors on validation images.  At test time, detections from multiple different detectors can be reconciled by fitting the distribution of detection scores of each detector to that of the Platt-scaled validation set.  After scaling, the maximum value of the combined detector vector ${\bf c}$ is used as the final fused score.  The WS approach finds weights of detection scores that maximize the product of a weight vector ${\bf w}$ and the detection vector of detector scores ${\bf c}$, $f_{WS}({\bf c}) = {\bf w}^T{\bf c}$.  ${\bf w}$ is learned through linear SVM optimization.  In WS, detection scores are converted into probabilities by Platt scaling as well because negative infinity scores in the combined detection score for the non-overlapping windows can hurt the SVM optimization.  For Bayesian fusion, we use a naive Bayesian model assuming that all the approaches are independent of each other.  In other words, the joint likelihood can be decomposed as the product of the likelihoods of each detector, while the posterior is expressed as the product of the prior and the joint likelihood (i.e. Bayes' rule).  The remaining two approaches, local expert forest (LEF)~\cite{JLiuECCV12} and Detect2Rank (D2R)~\cite{SKaraogluArxiv14}, have been recently introduced.  In~\cite{SKaraogluArxiv14}, Karaoglu et al. implemented four ranking approaches and we have used PoW2, the best among the four.  These current works are compared with our proposed algorithm only using PASCAL VOC07 dataset.

\begin{figure}[t]
    \begin{center}
    \includegraphics[trim = 10mm 30mm 10mm 25mm,width=0.46\textwidth]{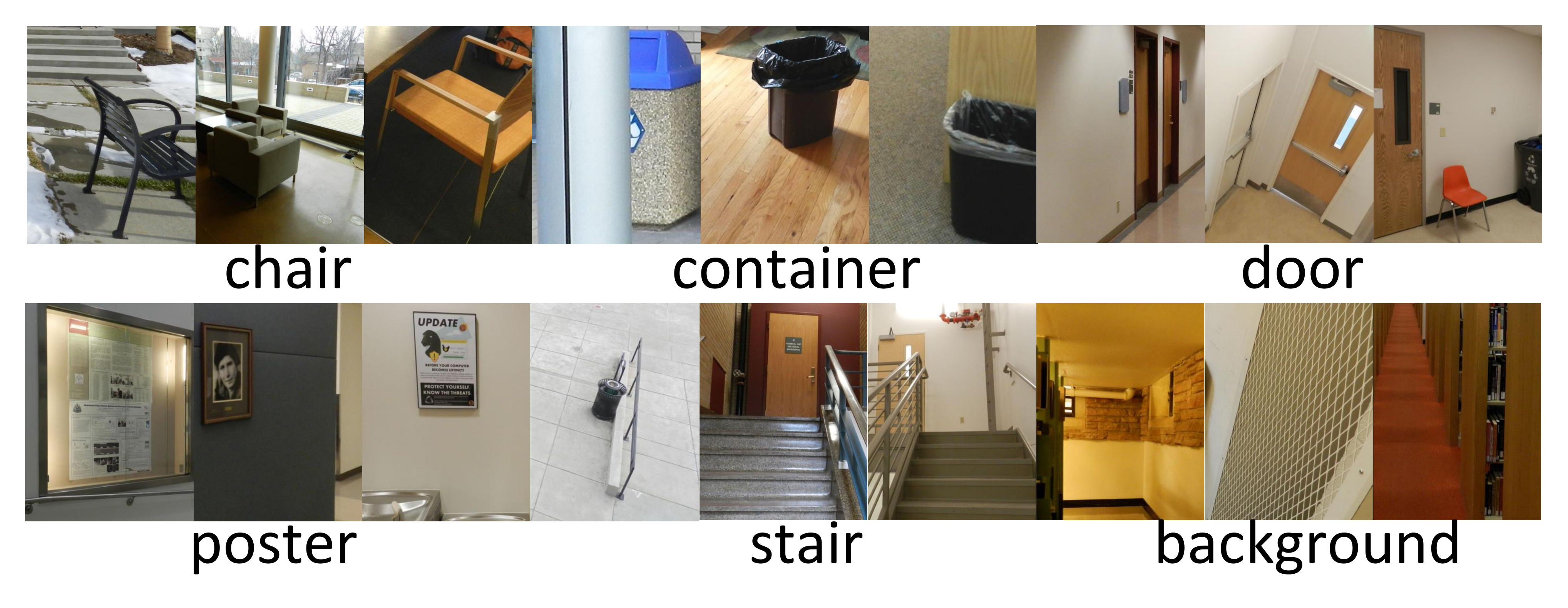}
    \end{center}
    \caption{ARL dataset~\cite{JTouryanNeuroprosthetics14}.}
    \label{fig:arl}
    \vspace{-0.20cm}
\end{figure}

To demonstrate the advantages of dynamic basic probability assignment in the proposed DBF, we also implement a regular DST fusion method that employs only static basic probability assignment~\cite{LXuSMC92}, in which each detector's pervious performance is characterized by the probabilities of the three hypotheses at the fixed precision value corresponding to a recall of 0.2.

\subsection{Evaluation of ARL Dataset}

The Army Research Lab (ARL) image dataset was originally created for the purpose of analyzing human performance in Rapid Serial Visual Presentation (RSVP)~\cite{JTouryanNeuroprosthetics14} tasks, but is also applicable to object detection tasks.  (In future work, we plan to integrate computer vision-based object detection with human decisions.)  The dataset contains 3000 images of both indoor and outdoor scenes, 1438 images of which contain at least one object-of-interest. The target objects include chair, container, door, poster, and stair.  Figure~\ref{fig:arl} displays several example images of all five objects as well as background images.  The number of images in the ARL dataset is relatively small compared to that of other benchmark datasets, such as PASCAL VOC 07 and ImageNet, but, with regard to the mean average precision (mAP), the ARL dataset (0.253 for DPM) is not considerably less challenging compared to the benchmark datasets (0.239 for DPM on PASCAL VOC 07).

The proposed DBF algorithm was evaluated on the ARL dataset and its average precision (AP) was compared to that of four individual detection algorithms (the HOG-SVM detector was not utilized on the ARL dataset.) and four other ``baseline'' fusion methods (Platt, Bayes, WS, and DST) for each object class.  Results are shown in Table~\ref{tab:map_arl}.  


\begin{table}[t]
\begin{center}
\begin{tabular}{l|ccccc|c}
\hline
& {\small chair} &{\small contr} & {\small door} & {\small postr} & {\small stair} & {\small mAP} \\\hline\hline
{\small DSIFT} & {\small .143} & {\small .037} & {\small .073} & {\small .143} & {\small .061} & {\small .091}\\\hline
{\small TAG} & {\small .045} & {\small .128} & {\small .165} & {\small .066} & {\small .008} & {\small .082}\\\hline
{\small ESVM} & {\small .125} & {\small .318} & {\small .150} & {\small .236} & {\small .122} & {\small .190}\\\hline
{\small DPM} & {\small .188} & {\small .396} & {\small .194} & {\small .342} & {\small .143} & {\small .253}\\\hline\hline
{\small Platt} & {\small .191} & {\small .364} & {\small .204} & {\small .307} & {\small .125} & {\small .238}\\\hline
{\small WS} & {\small .192} & {\small .388} & {\small .267} & {\small .318} & {\small .096} & {\small .252}\\\hline
{\small Bayes} & {\small .244} & {\small .424} & {\small .281} & {\small .341} & {\small .089} & {\small .276}\\\hline\hline
{\small DST} & {\small .234} & {\small .314} & {\small .230} & {\small .247} & {\small {\bf .168}} & {\small .238}\\\hline
{\small DBF} & {\small {\bf .329}} & {\small {\bf .451}} & {\small {\bf .298}} & {\small {\bf .390} } & {\small .159} & {\small {\bf .325}}\\\hline
\end{tabular}
\end{center}
\caption{AP on the ARL dataset~\cite{JTouryanNeuroprosthetics14}.}
\label{tab:map_arl}
\end{table}

\subsection{Evaluation of PASCAL VOC 07 Dataset}

\begin{table*}[t]
\begin{center}
\begin{tabular}{l|@{}c@{}c@{}c@{}c@{}c@{}c@{}c@{}c@{}c@{}c@{}c@{}c@{}c@{}c@{}c@{}c@{}c@{}c@{}c@{}c|@{}c}
\hline &{\scriptsize ~aero} &{\scriptsize ~bike} &{\scriptsize ~bird} &{\scriptsize ~boat} &{\scriptsize ~bottle} &{\scriptsize ~bus} &{\scriptsize ~car} &{\scriptsize ~cat} &{\scriptsize ~chair} &{\scriptsize ~cow} &{\scriptsize ~table} &{\scriptsize ~dog} &{\scriptsize ~horse} &{\scriptsize mbik} &{\scriptsize ~pers} &{\scriptsize ~plant} &{\scriptsize ~sheep} &{\scriptsize ~sofa} &{\scriptsize ~train} &{\scriptsize ~tv} &{\scriptsize ~~mAP} \\\hline\hline
{\scriptsize HOG} & {\scriptsize~~.036~~} & {\scriptsize~~.060~~} & {\scriptsize~~.001~~} & {\scriptsize~~.001~~} & {\scriptsize~~.005~~} & {\scriptsize~~.005~~} & {\scriptsize~~.094~~} & {\scriptsize~~.001~~} & {\scriptsize~~.001~~} & {\scriptsize~~.092~~} & {\scriptsize~~.001~~} & {\scriptsize~~.002~~} & {\scriptsize~~.002~~} & {\scriptsize~~.005~~} & {\scriptsize~~.001~~} & {\scriptsize~~.001~~} & {\scriptsize~~.003~~} & {\scriptsize~~.001~~} & {\scriptsize~~.013~~} & {\scriptsize~~.103~~} & {\scriptsize ~~.021} \\\hline
{\scriptsize TAG} & {\scriptsize~~.019~~} & {\scriptsize~~.051~~} & {\scriptsize~~.009~~} & {\scriptsize~~.002~~} & {\scriptsize~~.002~~} & {\scriptsize~~.028~~} & {\scriptsize~~.022~~} & {\scriptsize~~.080~~} & {\scriptsize~~.002~~} & {\scriptsize~~.006~~} & {\scriptsize~~.056~~} & {\scriptsize~~.032~~} & {\scriptsize~~.020~~} & {\scriptsize~~.085~~} & {\scriptsize~~.051~~} & {\scriptsize~~.002~~} & {\scriptsize~~.001~~} & {\scriptsize~~.010~~} & {\scriptsize~~.020~~} & {\scriptsize~~.014~~} & {\scriptsize ~~.026} \\\hline
{\scriptsize DSIFT} & {\scriptsize~~.081~~} & {\scriptsize~~.024~~} & {\scriptsize~~.017~~} & {\scriptsize~~.004~~} & {\scriptsize~~.002~~} & {\scriptsize~~.080~~} & {\scriptsize~~.118~~} & {\scriptsize~~.142~~} & {\scriptsize~~.005~~} & {\scriptsize~~.097~~} & {\scriptsize~~.109~~} & {\scriptsize~~.128~~} & {\scriptsize~~.040~~} & {\scriptsize~~.037~~} & {\scriptsize~~.076~~} & {\scriptsize~~.002~~} & {\scriptsize~~.059~~} & {\scriptsize~~.102~~} & {\scriptsize~~.122~~} & {\scriptsize~~.028~~} & {\scriptsize ~~.064} \\\hline
{\scriptsize ESVM} & {\scriptsize~~.164~~} & {\scriptsize~~.418~~} & {\scriptsize~~.041~~} & {\scriptsize~~.096~~} & {\scriptsize~~.107~~} & {\scriptsize~~.341~~} & {\scriptsize~~.336~~} & {\scriptsize~~.095~~} & {\scriptsize~~.100~~} & {\scriptsize~~.129~~} & {\scriptsize~~.097~~} & {\scriptsize~~.013~~} & {\scriptsize~~.362~~} & {\scriptsize~~.322~~} & {\scriptsize~~.170~~} & {\scriptsize~~.033~~} & {\scriptsize~~.170~~} & {\scriptsize~~.102~~} & {\scriptsize~~.287~~} & {\scriptsize~~.263~~} & {\scriptsize ~~.182} \\\hline

{\scriptsize Color attributes} & {\scriptsize~~.0.201~~} & {\scriptsize~~.518~~} & {\scriptsize~~.026  ~~} & {\scriptsize~~.102~~} & {\scriptsize~~.167~~} & {\scriptsize~~.344~~} & {\scriptsize~~.363~~} & {\scriptsize~~.172~~} & {\scriptsize~~.158~~} & {\scriptsize~~.198~~} & {\scriptsize~~.041~~} & {\scriptsize~~.358~~} & {\scriptsize~~.349~~} & {\scriptsize~~.436~~} & {\scriptsize~~.376~~} & {\scriptsize~~.106~~} & {\scriptsize~~.128~~} & {\scriptsize~~.273~~} & {\scriptsize~~.304~~} & {\scriptsize~~.307~~} & {\scriptsize ~~.246} \\\hline

{\scriptsize DPM} & {\scriptsize~~.231~~} & {\scriptsize~~.500~~} & {\scriptsize~~.036~~} & {\scriptsize~~.099~~} & {\scriptsize~~ .162~~} & {\scriptsize~~ .388~~} & {\scriptsize~~.451~~} & {\scriptsize~~.153~~} & {\scriptsize~~.120~~} & {\scriptsize~~.172~~} & {\scriptsize~~.129~~} & {\scriptsize~~.106~~} & {\scriptsize~~.463~~} & {\scriptsize~~.375~~} & {\scriptsize~~.346~~} & {\scriptsize~~.109~~} & {\scriptsize~~.109~~} & {\scriptsize~~.144~~} & {\scriptsize~~.353~~} & {\scriptsize~~{\bf .333}~~} & {\scriptsize ~~.239} \\\hline

{\scriptsize CNN} & {\scriptsize~~.010~~} & {\scriptsize~~.080~~} & {\scriptsize~~.035~~} & {\scriptsize~~.031~~} & {\scriptsize~~.001~~} & {\scriptsize~~.048~~} & {\scriptsize~~.030~~} & {\scriptsize~~.074~~} & {\scriptsize~~.011~~} & {\scriptsize~~.040~~} & {\scriptsize~~.039~~} & {\scriptsize~~.063~~} & {\scriptsize~~.099~~} & {\scriptsize~~.078~~} & {\scriptsize~~.035~~} & {\scriptsize~~.022~~} & {\scriptsize~~.022~~} & {\scriptsize~~.018~~} & {\scriptsize~~.046~~} & {\scriptsize~~.034~~} & {\scriptsize ~~.041}\footnotemark \\\hline
{\scriptsize RCNN} & {\scriptsize~~.637~~} & {\scriptsize~~.709~~} & {\scriptsize~~{\bf.506}~~} & {\scriptsize~~{\bf.393}~~} & {\scriptsize~~.300~~} & {\scriptsize~~.639~~} & {\scriptsize~~.721~~} & {\scriptsize~~{\bf.601}~~} & {\scriptsize~~.303~~} & {\scriptsize~~{\bf.585}~~} & {\scriptsize~~.458~~} & {\scriptsize~~{\bf.559}~~} & {\scriptsize~~.631~~} & {\scriptsize~~{\bf.681}~~} & {\scriptsize~~.549~~} & {\scriptsize~~.291~~} & {\scriptsize~~.536~~} & {\scriptsize~~.467~~} & {\scriptsize~~.575~~} & {\scriptsize~~.662~~} & {\scriptsize ~~.540} \\\hline\hline

{\scriptsize Platt} & {\scriptsize~~.596~~} & {\scriptsize~~.695~~} & {\scriptsize~~.470~~} & {\scriptsize~~.383~~} & {\scriptsize~~.314~~} & {\scriptsize~~.627~~} & {\scriptsize~~.708~~} & {\scriptsize~~.566~~} & {\scriptsize~~.295~~} & {\scriptsize~~.542~~} & {\scriptsize~~.398~~} & {\scriptsize~~.529~~} & {\scriptsize~~.595~~} & {\scriptsize~~.640~~} & {\scriptsize~~.508~~} & {\scriptsize~~.278~~} & {\scriptsize~~.503~~} & {\scriptsize~~.439~~} & {\scriptsize~~.537~~} & {\scriptsize~~.605~~} & {\scriptsize ~~.511} \\\hline
{\scriptsize WS} & {\scriptsize~~.576~~} & {\scriptsize~~.692~~} & {\scriptsize~~.486~~} & {\scriptsize~~.370~~} & {\scriptsize~~.326~~} & {\scriptsize~~.601~~} & {\scriptsize~~.706~~} & {\scriptsize~~.526~~} & {\scriptsize~~.315~~} & {\scriptsize~~.533~~} & {\scriptsize~~.450~~} & {\scriptsize~~.511~~} & {\scriptsize~~.658~~} & {\scriptsize~~.628~~} & {\scriptsize~~.538~~} & {\scriptsize~~.273~~} & {\scriptsize~~.502~~} & {\scriptsize~~.466~~} & {\scriptsize~~.577~~} & {\scriptsize~~.594~~} & {\scriptsize ~~.516} \\\hline

{\scriptsize LEF} & {\scriptsize~~.606~~} & {\scriptsize~~.671~~} & {\scriptsize~~.441~~} & {\scriptsize~~.366~~} & {\scriptsize~~.291~~} & {\scriptsize~~.624~~} & {\scriptsize~~.721~~} & {\scriptsize~~.503~~} & {\scriptsize~~.300~~} & {\scriptsize~~.571~~} & {\scriptsize~~.444~~} & {\scriptsize~~.463~~} & {\scriptsize~~.621~~} & {\scriptsize~~.615~~} & {\scriptsize~~.524~~} & {\scriptsize~~.276~~} & {\scriptsize~~.503~~} & {\scriptsize~~.488~~} & {\scriptsize~~.528 ~~} & {\scriptsize~~.628~~} & {\scriptsize ~~.510} \\\hline
{\scriptsize D2R} & {\scriptsize~~.609~~} & {\scriptsize~~.687~~} & {\scriptsize~~.468~~} & {\scriptsize~~.398~~} & {\scriptsize~~.311~~} & {\scriptsize~~{\bf.665}~~} & {\scriptsize~~{\bf .757}~~} & {\scriptsize~~.552~~} & {\scriptsize~~.326~~} & {\scriptsize~~.587~~} & {\scriptsize~~.449~~} & {\scriptsize~~.493~~} & {\scriptsize~~.660~~} & {\scriptsize~~.636~~} & {\scriptsize~~.528~~} & {\scriptsize~~.289~~} & {\scriptsize~~.511~~} & {\scriptsize~~.502~~} & {\scriptsize~~.550 ~~} & {\scriptsize~~.654~~} & {\scriptsize ~~.531} \\\hline

{\scriptsize Bayes} & {\scriptsize~~.460~~} & {\scriptsize~~.616~~} & {\scriptsize~~.177~~} & {\scriptsize~~.098~~} & {\scriptsize~~.297~~} & {\scriptsize~~.541~~} & {\scriptsize~~.644~~} & {\scriptsize~~.252~~} & {\scriptsize~~.115~~} & {\scriptsize~~.413~~} & {\scriptsize~~.278~~} & {\scriptsize~~.344~~} & {\scriptsize~~.359~~} & {\scriptsize~~.517~~} & {\scriptsize~~.229~~} & {\scriptsize~~.215~~} & {\scriptsize~~.447~~} & {\scriptsize~~.138~~} & {\scriptsize~~.461~~} & {\scriptsize~~.475~~} & {\scriptsize ~~.354} \\\hline\hline
{\scriptsize DST} & {\scriptsize~~.423~~} & {\scriptsize~~.601~~} & {\scriptsize~~.291~~} & {\scriptsize~~.250~~} & {\scriptsize~~.259~~} & {\scriptsize~~.580~~} & {\scriptsize~~.624~~} & {\scriptsize~~.382~~} & {\scriptsize~~.215~~} & {\scriptsize~~.348~~} & {\scriptsize~~.352~~} & {\scriptsize~~.291~~} & {\scriptsize~~.532~~} & {\scriptsize~~.511~~} & {\scriptsize~~.483~~} & {\scriptsize~~.227~~} & {\scriptsize~~.345~~} & {\scriptsize~~.277~~} & {\scriptsize~~.456~~} & {\scriptsize~~.488~~} & {\scriptsize ~~.397} \\\hline
{\scriptsize DBF} & {\scriptsize~~{\bf .650}~~} & {\scriptsize~~{\bf .720}~~} & {\scriptsize~~.501~~} & {\scriptsize~~.392~~} & {\scriptsize~~{\bf .341}~~} & {\scriptsize~~ .658~~} & {\scriptsize~~.729~~} & {\scriptsize~~.576~~} & {\scriptsize~~{\bf .339}~~} & {\scriptsize~~.578~~} & {\scriptsize~~{\bf .477}~~} & {\scriptsize~~.537~~} & {\scriptsize~~{\bf.670}~~} & {\scriptsize~~.664~~} & {\scriptsize~~{\bf .572}~~} & {\scriptsize~~{\bf .315}~~} & {\scriptsize~~{\bf .537}~~} & {\scriptsize~~{\bf .539}~~} & {\scriptsize~~{\bf.590}~~} & {\scriptsize~~{\bf.672}~~} & {\scriptsize ~~{\bf .553}} \\\hline
\end{tabular}
\end{center}
\caption{AP on the PASCAL VOC 07 dataset~\cite{pascal-voc-2007}.}

\label{tab:map_pascal}
\end{table*}

The fusion and individual detection methods were also evaluated on the PASCAL VOC 07 dataset~\cite{pascal-voc-2007}.  PASCAL VOC 07 provides {\tt train}, {\tt val}, {\tt trainval}, and {\tt test}, where the {\tt trainval} set consists of images of {\tt train} and {\tt val} sets.  While previous works that used individual detectors employed in our fusion method use the {\tt trainval} set , we learn the detectors and trust models on {\tt train} and {\tt val} set, respectively.  This split is made to avoid building trust models that overfit the training dataset.  Therefore, the performance of the individual detectors used in our work is worse than the performance reported in the original literature with regard to the individual detectors, as we are using a smaller training dataset.


The mAP of each individual detector and fusion method is reported in Table~\ref{tab:map_pascal}.  To evaluate fusion on the PASCAL VOC 07 dataset, eight individual detectors (DSIFT-SVM, HOG-SVM, TAG, Exemplar SVM, two DPMs employing HOG and color attributes, separately, fine-tuned CNN (ft-CNN) and RCNN) were selected and fusion of their detection results was conducted.  

\section{Discussion}
\label{sec:discuss}

Both mAP and ROC performance metrics show that DBF outperformed all the baseline fusion algorithms as well as individual detectors on both ARL and PASCAL VOC 07 datasets. DBF demonstrates the best results for 4 of 5 categories in the ARL dataset and 12 of 20 categories in the PASCAL VOC 07 dataset. Notably DBF is the only fusion approaches that outperforms RCNN on the PASCAL VOC 07 though improvement is small.  Only a minor improvement is achieved because the performance of RCNN is much greater than the other detectors.  Therefore, we evaluated fusion performance again, but without RCNN, and the results are present in table~\ref{tab:map_pascal2}.  In table~\ref{tab:map_pascal2}, DBF still outperformed all baseline fusion methods and all individual detectors with a significant gain in mAP (.06 from LEF and .10 from color attributes).  The clear difference in performance between conventional DST fusion and the proposed DBF demonstrates the strength of dynamic basic probability assignment over the conventional method of static assignment.  Likewise, the fact that DBF outperforms Bayesian fusion demonstrates the benefits of adding an intermediate state to the set of hypotheses.

\begin{table}[t]
\begin{center}
\begin{tabular}{c|c|c|c|c|c|c|c}
\hline
 & Platt & WS & LEF & D2R & Bayes & DST & DBF \\\hline\hline
mAP & .268 & .271 &  .283 & .261 & .253 & .257 & {\bf.341} \\\hline
\end{tabular}
\end{center}
\caption{Performance of fusion approaches with all the detectors except RCNN.}
\label{tab:map_pascal2}
\end{table}

\begin{figure*}[t]
    \begin{center}
    \begin{subfigure}[t]{0.33\textwidth}
	\begin{center}
    	\includegraphics[trim = 0mm 130mm 100mm 0mm,clip=true,width=\textwidth]{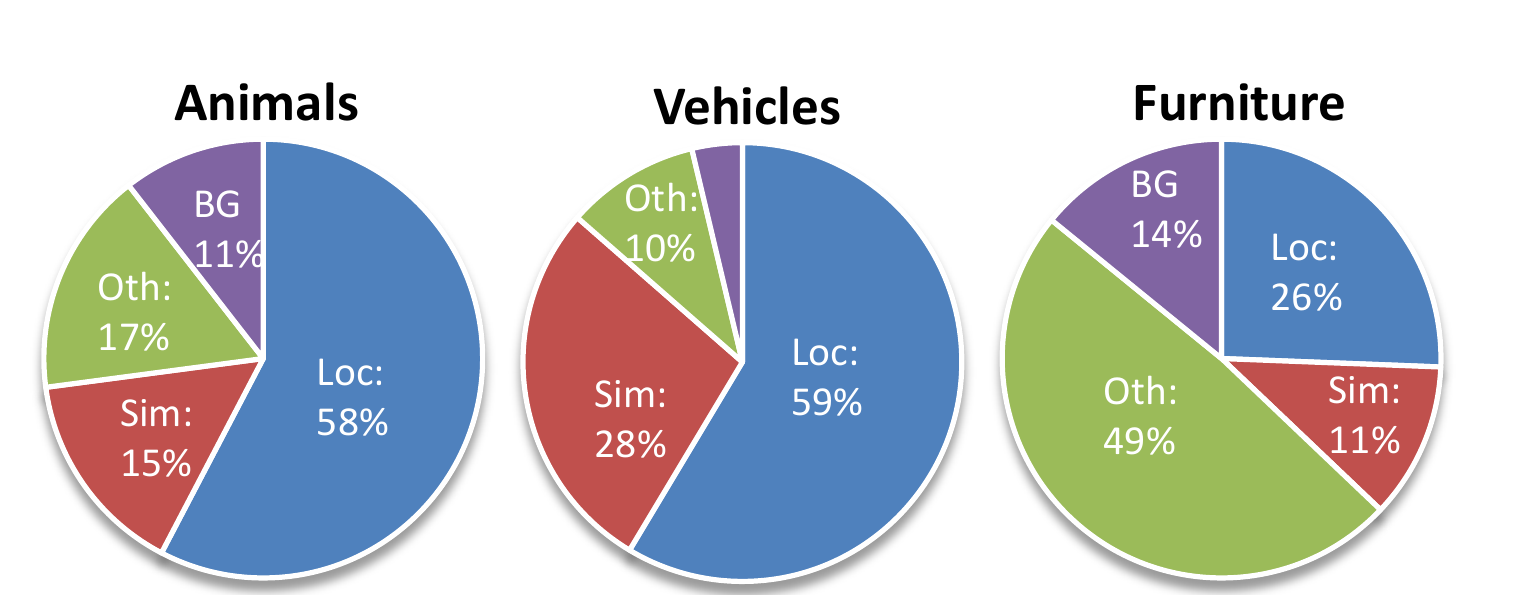}
	\end{center}
	\caption{DSIFT}
    \end{subfigure}
    \begin{subfigure}[t]{0.33\textwidth}
	\begin{center}
    	\includegraphics[trim = 0mm 130mm 100mm 0mm,clip=true,width=\textwidth]{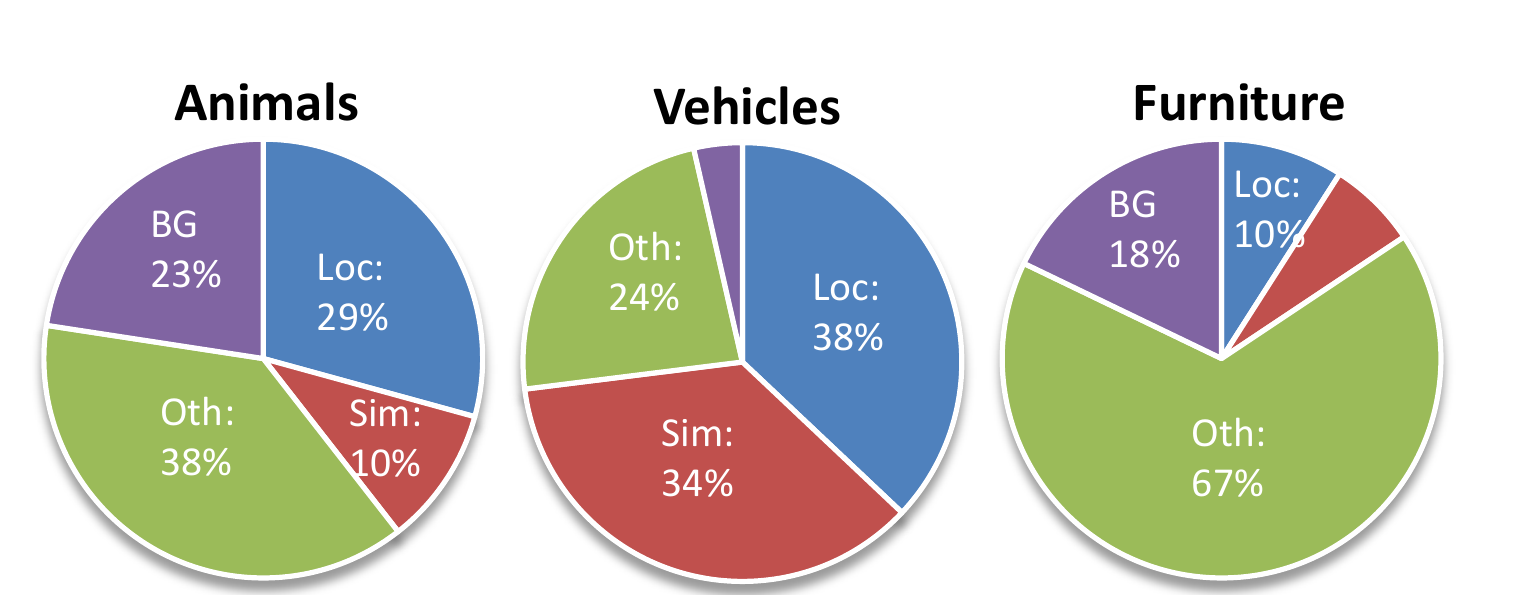}
	\end{center}
	\caption{TAG}
    \end{subfigure}
    \begin{subfigure}[t]{0.33\textwidth}
	\begin{center}
    	\includegraphics[trim = 0mm 130mm 100mm 0mm,clip=true,width=\textwidth]{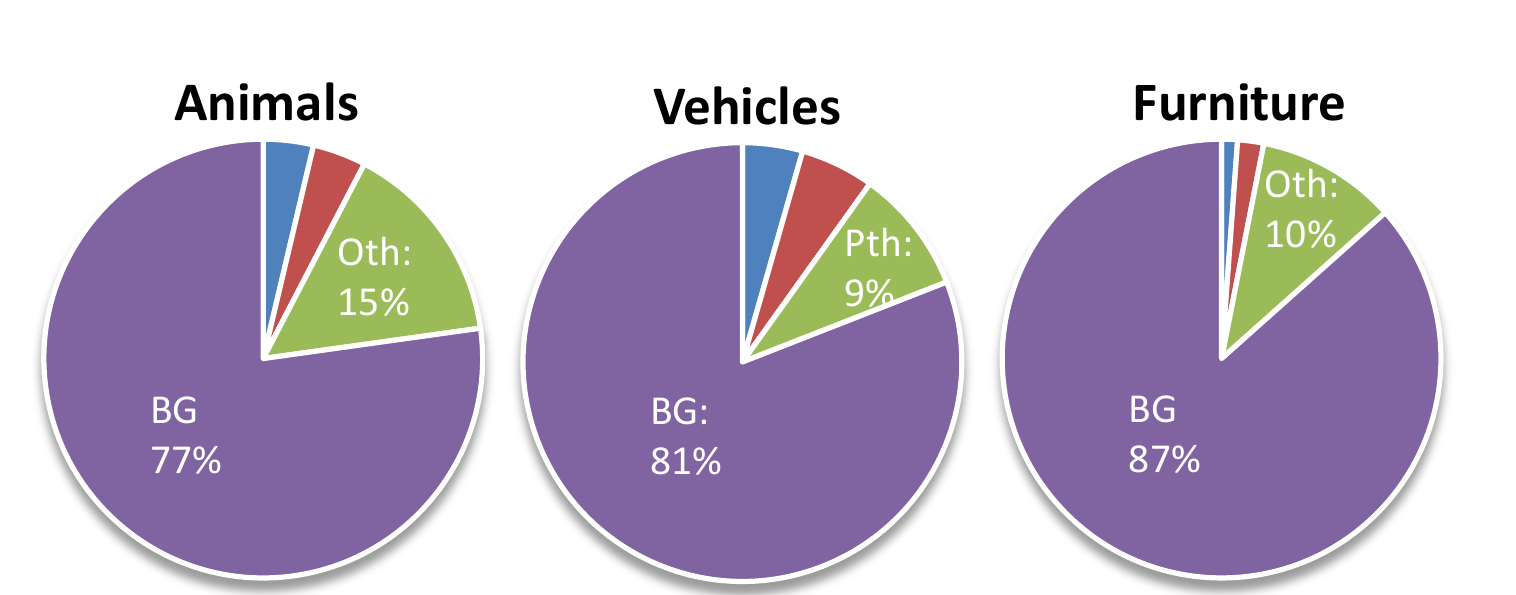}
	\end{center}
	\caption{PHOG}
    \end{subfigure} \\
    \begin{subfigure}[t]{0.33\textwidth}
	\begin{center}
    	\includegraphics[trim = 0mm 130mm 100mm 0mm,clip=true,width=\textwidth]{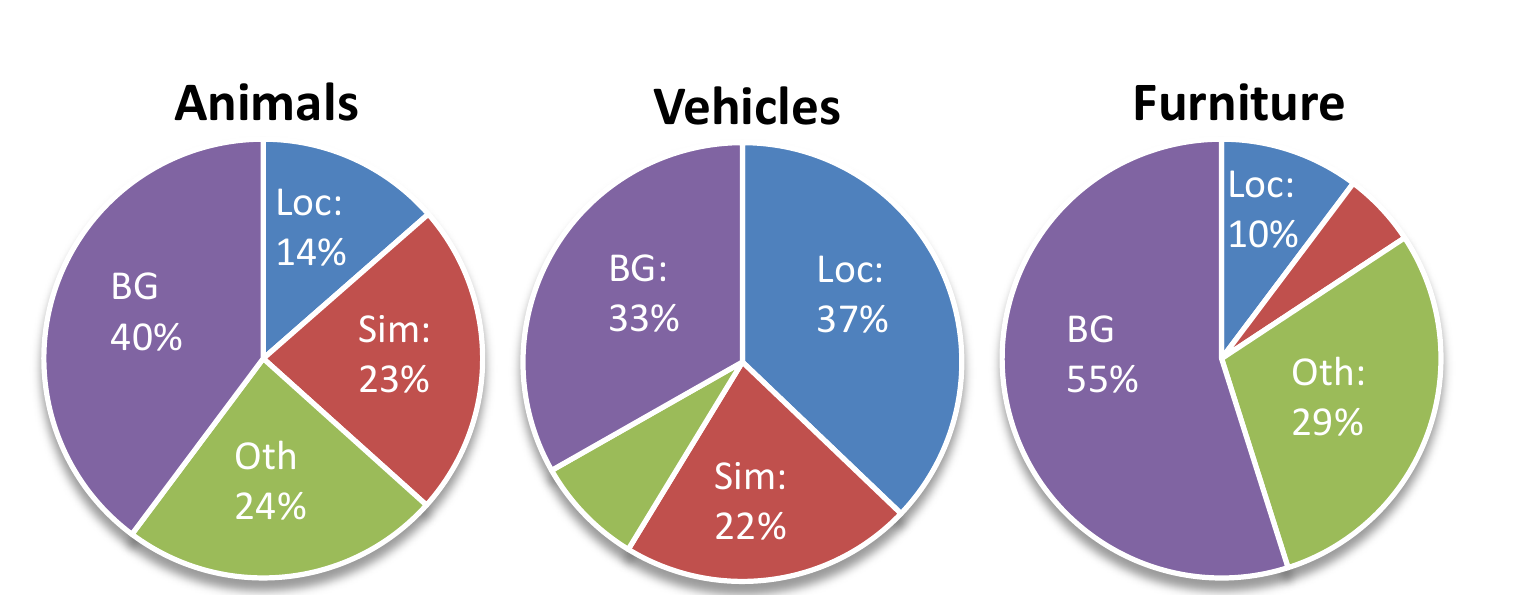}
	\end{center}
	\caption{Exemplar SVM}
    \end{subfigure}
    \begin{subfigure}[t]{0.33\textwidth}
	\begin{center}
    	\includegraphics[trim = 0mm 130mm 100mm 0mm,clip=true,width=\textwidth]{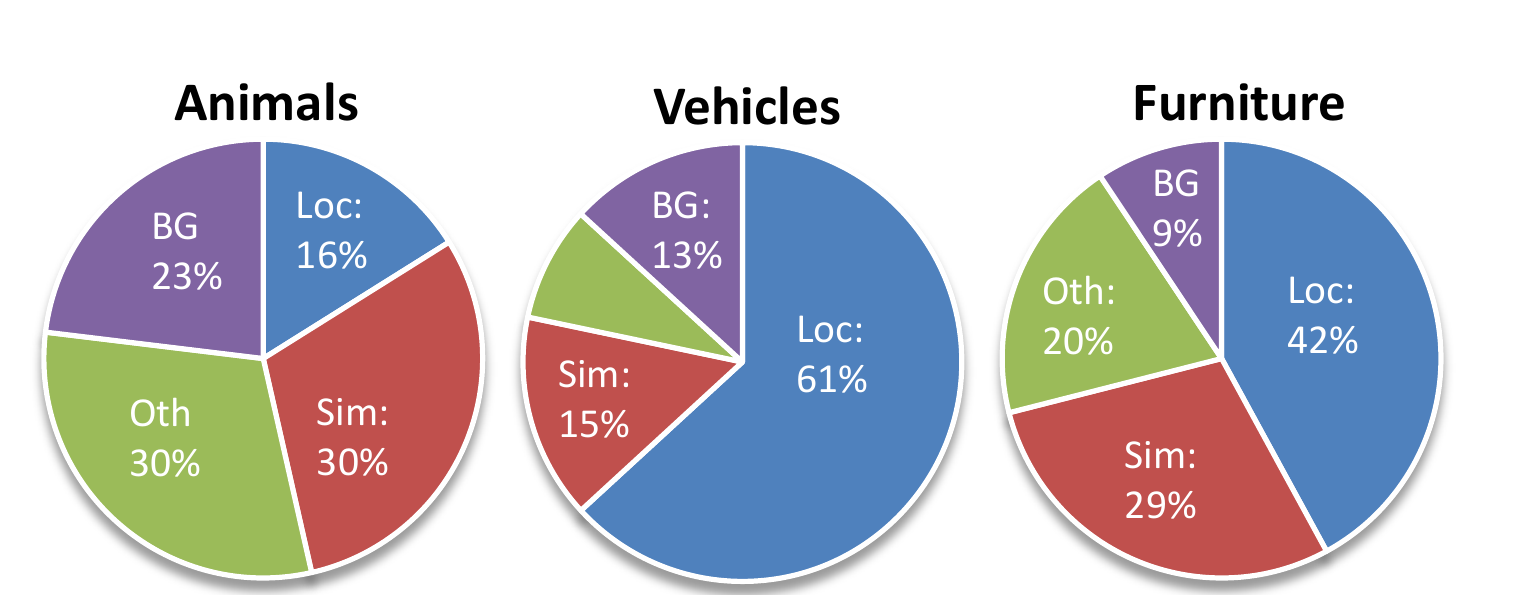}
	\end{center}
	\caption{Color Attrubutes}
    \end{subfigure}
    \begin{subfigure}[t]{0.33\textwidth}
	\begin{center}
    	\includegraphics[trim = 0mm 130mm 100mm 0mm,clip=true,width=\textwidth]{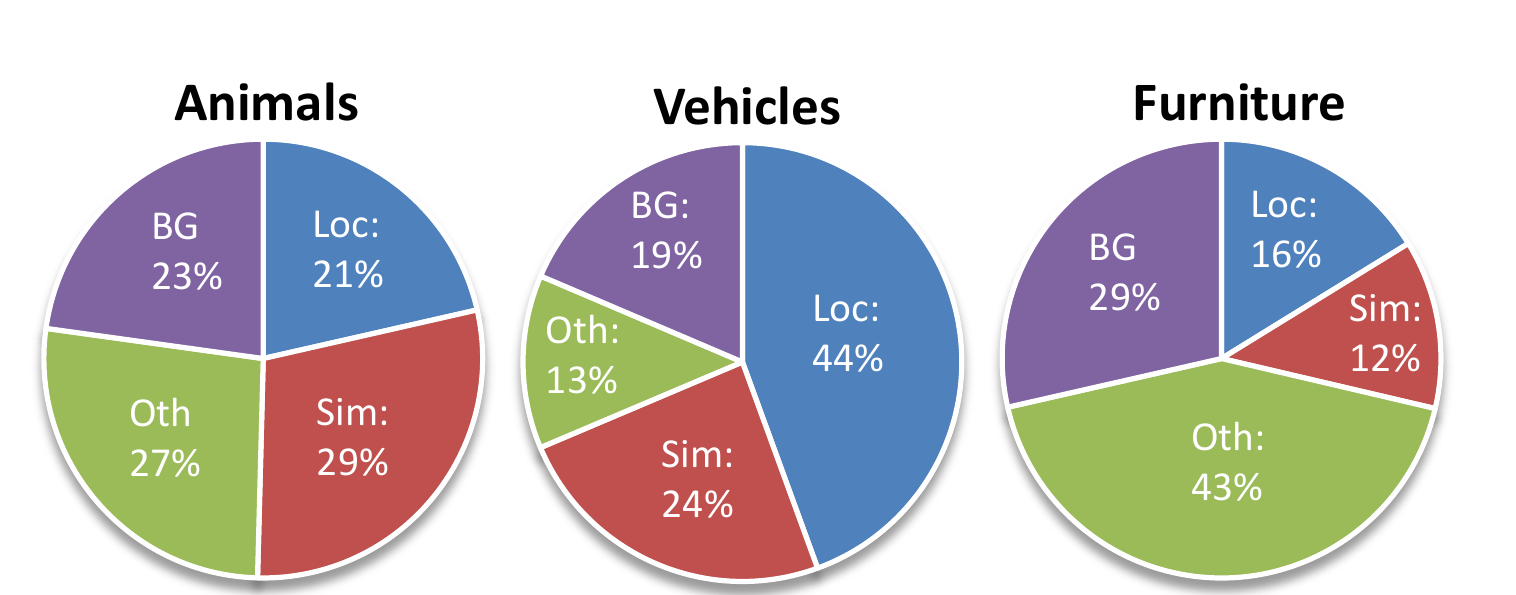}
	\end{center}
	\caption{DPM}
    \end{subfigure} \\
    \begin{subfigure}[t]{0.33\textwidth}
	\begin{center}
    	\includegraphics[trim = 0mm 130mm 100mm 0mm,clip=true,width=\textwidth]{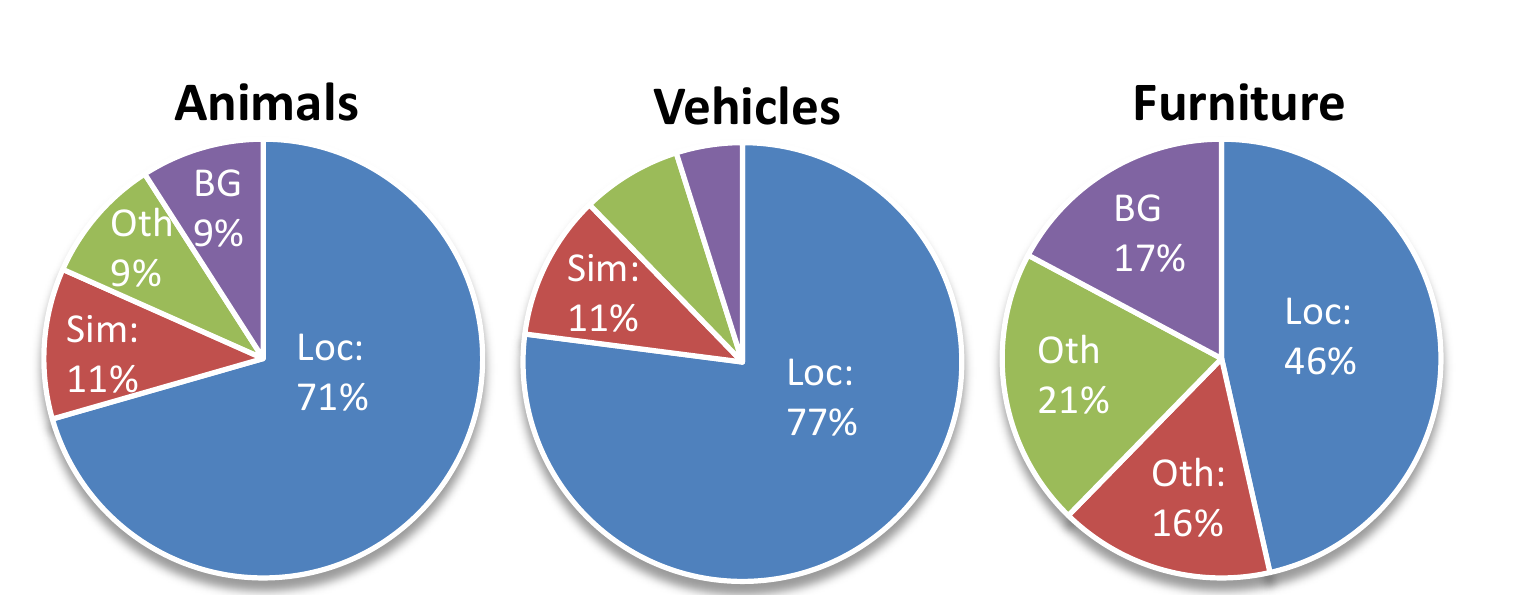}
	\end{center}
	\caption{CNN}
    \end{subfigure}
    \begin{subfigure}[t]{0.33\textwidth}
	\begin{center}
    	\includegraphics[trim = 0mm 130mm 100mm 0mm,clip=true,width=\textwidth]{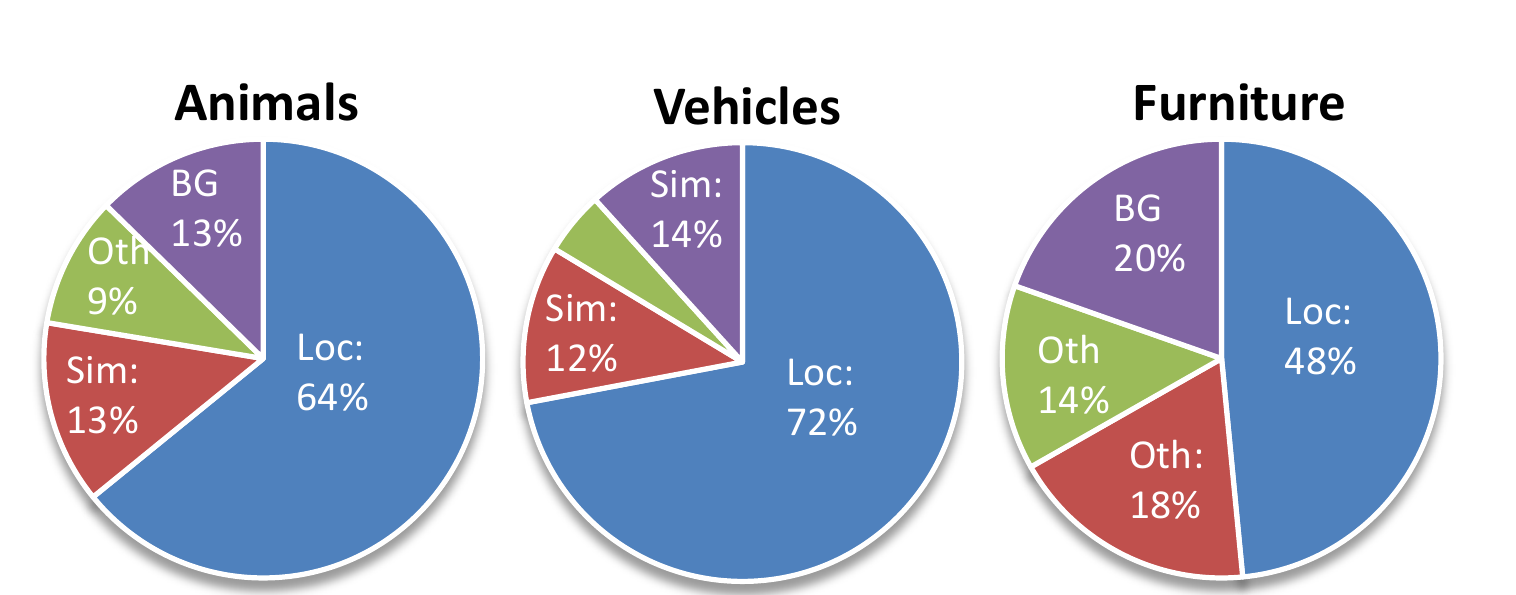}
	\end{center}
	\caption{RCNN}
    \end{subfigure}
    \begin{subfigure}[t]{0.33\textwidth}
	\begin{center}
    	\includegraphics[trim = 0mm 130mm 100mm 0mm,clip=true,width=\textwidth]{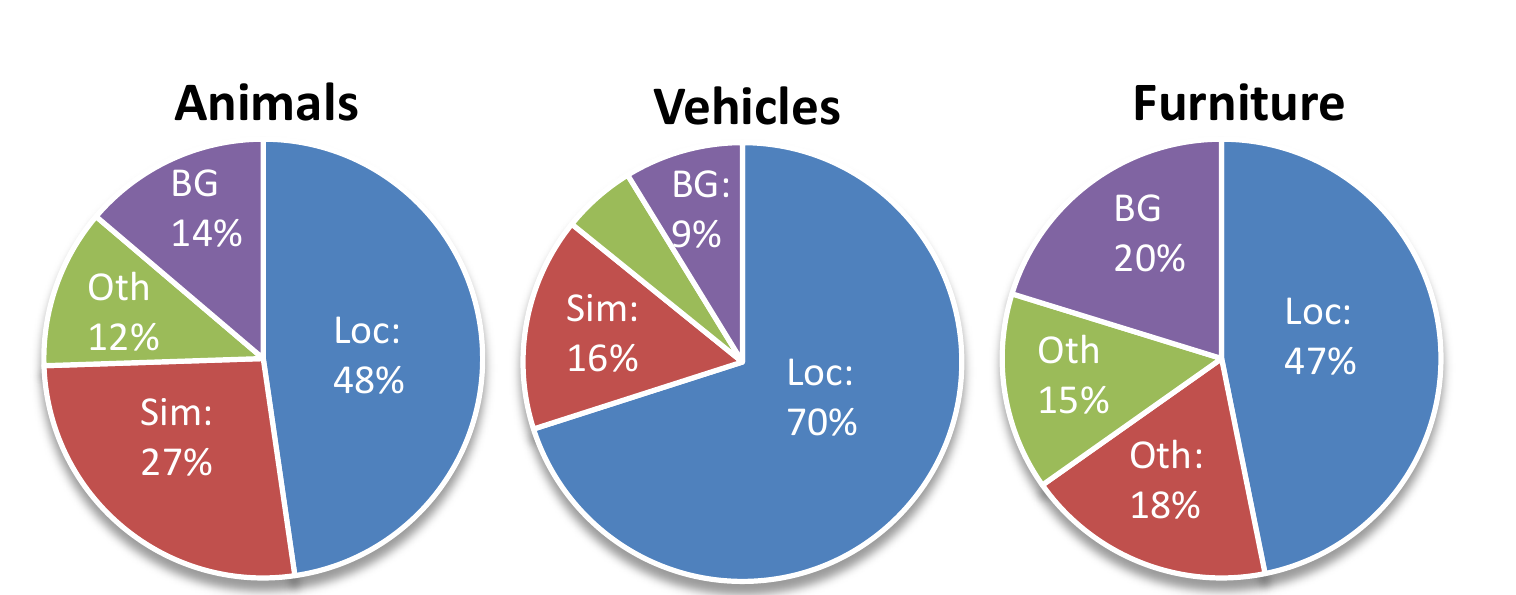}
	\end{center}
	\caption{DBF}
    \end{subfigure} \\
    \end{center}
    \caption{{\bf Analysis of top-ranked false positives:} Pie charts present the fractions of four types of top-ranked false positives.  Analysis is performed on PASCAL VOC 07 dataset.  Among 20 object categories in PASCAL VOC 07 dataset, all animals including person are in ``Animal'', all vehicles are in ``Vehicle'', and `chair', `diningtable', and `sofa' are assigned to ``Furniture''.  Localization error (Loc), confusion with similar classes (Sim), confusion with dissimilar object categories (Oth), and confusion with background (BG) are indicated by blu, red, green, and purple face, respectively.}

    \label{fig:diag_fp}
\end{figure*}

In addition, we analyzed top-ranked false positives on the PASCAL VOC 07 and categorize them into 4 types according to Hoiem et al.~\cite{DHoiemECCV12} in Figure~\ref{fig:diag_fp}.  Four types of false positives (FP) are poor localization (Loc), confusion with similar classes (Sim), Confusion with dissimilar object categories (Oth), and confusion with background (BG).  Notably, most of FP in CNN performance is from poor localization.  We can guess that CNN performed much worse than expectation because coarse-grid scanning windows and aspect ratio of windows fixed as square bring localization error.  FPs detected by RCNN have similar fractions to CNN. Once accurate localization approaches replaces coarse-grid sliding windows (employed by CNN) or objectness (employed by RCNN), CNN-based detector may achieve much better performance. The charts also demonstrates that, as compared with RCNN, DBF increases the performance by reducing inaccurately localized false positives.

\begin{table}[t]
\begin{center}
\begin{tabular}{c|c|c}
\hline
\# of detectors & ARL & PASCAL \\\hline\hline
2 & .295 & .545\\
3 & .319 & .547\\
4 & .325 & .548\\
5 &  & .548\\
6 && .552\\
7 && .553\\ 
8 && .553\\\hline
\end{tabular}
\end{center}
\caption{Comparison of fusion performance with respect to the combination of multiple detectors.}
\label{tab:eval_best_comb}
\end{table}

In order to further investigate whether (and to what degree) complementary information is provided by each detector using DBF, mAP was calculated while varying the number of individual detectors used in fusion. For each combination number $K$, detectors with the $K$ highest mAP were selected. Results, shown in Table~\ref{tab:eval_best_comb} for both the ARL and PASCAL datasets, illustrate that performance improves as the number of detectors increases, at a decreasing rate. Note that the final row corresponds to the maximum number of combined detectors (4 for ARL, 5 for PASCAL).

\begin{figure}[t]
    \begin{center}
    \includegraphics[trim = 20mm 107mm 30mm 107mm,clip=true,width=0.45\textwidth]{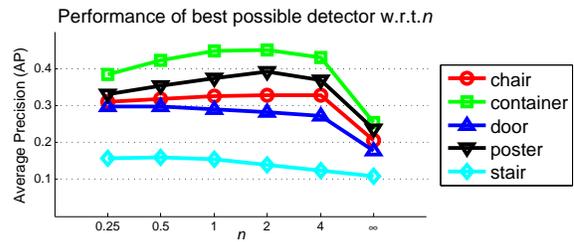}
    \end{center}
    \vspace{-0.3cm}
    \caption{Comparison of fusion performance with respect to the various theoretical \emph{best possible} detectors. $n$ in $x$ axis is the exponent in Eq.~\ref{eq:fn_bpd}.}
    \label{fig:map_ideal_detector}
\end{figure}

\footnotetext{Unexpected low performance for CNN is due to coarse-grid scanning window strategy with fixed square aspect ratio.  In section~\ref{sec:discuss}, further analysis about CNN performance is provided.}

Figure~\ref{fig:map_ideal_detector} illustrates the variation in mAP as the shape of the PR curve of the best possible detector is varied, for each object category in the ARL dataset. The optimal value of the parameter n (see Eq. 4), which dictates the shape (and hence, estimated performance) of the best possible detector, is different for different object categories. However, note that the notional perfect detector ($n$ = $\infty$) underperforms other choices of $n$ in every object category. This result suggests that our method of splitting the false positives into \emph{non-target} and \emph{intermediate state} categories is actually beneficial.

\section{Conclusions}
\label{sec:concl}

A novel fusion method, referred to as Dynamic Belief Fusion (DBF), is proposed to improve upon current late fusion methods in the context of object detection.  DBF employs prior information in the form of dynamic basic probability assignments. For object detection, these dynamic basic probability assignments (\emph{target}, \emph{non-target}, and \emph{intermediate state}) are generated from the precision-recall curve of a validation image set. In order to properly separate the \emph{non-target} and \emph{intermediate states}, the concept of a best possible detector is introduced and applied. Dempster's combination rule is used to combine the resulting basic probabilities of detections from different detectors.

Experimental results on two datasets ARL and PASCAL VOC 07, demonstrate that DBF outperforms all baseline fusion approaches as well as all individual detectors in terms of mean average precision (mAP).  DBF also achieved performance improvement over RCNN on PASCAL VOC 07.  Its superior performance as compared to the DST-based fusion approach (incorporating fixed levels of basic probabilities) clearly illustrates the robustness of dynamic basic probability assignment. Enhanced performance over Bayesian fusion supports the use of an intermediate belief state, which was achieved in this context via the instantiation of a best possible detector.

Also, note that DBF is a novel approach guaranteed to provide improved fusion performance over the best detector in conjunction with other detectors in the fusion pool through dynamic belief assignments and the Dempster-Shafer combination of assigned probabilities.  Therefore, addition and removal of individual detectors from the fusion pool can only further improve fusion performance as state-of-the-art detectors, such as deep learning approaches, are introduced.

{\small
\bibliographystyle{ieee}
\bibliography{egbib}
}

\end{document}